\documentclass{article}

\usepackage{times}
\usepackage{graphicx} 
\usepackage{subfigure} 

\usepackage{natbib}

\usepackage{algorithm}
\usepackage{algorithmic}

\usepackage{amsmath,amssymb} 
\usepackage{natbib}
\usepackage{hyperref}
\usepackage{color}

\usepackage{algorithm}
\usepackage{algorithmic}
\usepackage{qtree}

\usepackage{hyperref}

\usepackage[accepted]{icml2014}

\begin{document}

\newcommand{\todo}[1]{}
\newtheorem{thm}{Theorem}
\newtheorem{definition}{Definition}
\newtheorem{proposition}{Proposition}

\twocolumn[
\icmltitle{A reversible infinite HMM using normalised random measures}

\icmlauthor{Konstantina Palla}{kp376@cam.ac.uk}
\icmladdress{University of Cambridge}
\icmlauthor{David A. Knowles}{davidknowles@cs.stanford.edu}
\icmladdress{University of Cambridge}
\icmlauthor{Zoubin Ghahramani}{zoubin@eng.cam.ac.uk}
\icmladdress{University of Cambridge}


\vskip 0.3in
]

\begin{abstract}
We present a nonparametric prior over reversible Markov chains. We use completely random measures, specifically gamma processes, to construct a countably infinite graph with weighted edges.  By enforcing symmetry to make the edges undirected we define a prior over random walks on graphs that results in a reversible Markov chain. The resulting prior over infinite transition matrices is closely related to the hierarchical Dirichlet process but enforces reversibility. A reinforcement scheme has recently been proposed with similar properties, but the de Finetti measure is not well characterised. We take the alternative approach of explicitly constructing the mixing measure, which allows more straightforward and efficient inference at the cost of no longer having a closed form predictive distribution. We use our process to construct a reversible infinite HMM which we apply to two real datasets, one from epigenomics and one ion channel recording. 
\end{abstract}

\section{Introduction}
Consider a sequence of states $X_1, \dots, X_T$ sampled from a reversible Markov chain. A Markov chain is said to be \textit{reversible} if the probability of the chain is the same observed either forwards or backwards in time. Reversibility is a realistic assumption in various settings. For instance, reversible Markov chains are appropriate to model the time-reversal dynamics in physical systems, such as the transitions of a macromolecule conformation at fixed temperature or chemical dynamics in protein folding. In these settings, the system transitions between hidden states over time emitting a sequence of observations $Y_1, \dots, Y_T$. Our aim is to recover the process and hidden state sequence that gave rise to this observed data. To do this we define a prior over reversible Markov chains.

There is a close connection between reversible Markov chains and random walks on graphs. More specifically, a random walk on a weighted undirected graph produces a reversible Markov chain. In a random walk on a graph, the traveller jumps to the next node (state) with probability propotional to the corresponding edge weight. The aim now is to put a prior over the unknown transition matrix (analogously the weights) that guides the walk. Much research has gone into connecting random walks on graphs to reversible Markov chains with seminal works by \citet{diaconis80} and \citet{DiacCop86}. The latter assumes an edge reinforcement random walk (ERRW) where the edge weight is increased by one each time an edge is crossed. The process is defined for a finite state space and, in the limit, gives weights that are distributed according to an explicitly characterised mixing measure, which can be a conjugate prior for the reinforcement process. In the more recent work of \citet{bacallado2013}, the authors define a three-
parameter random walk with reinforcement, named the $(\theta, \alpha, \beta)$ scheme, which generalizes the linearly edge reinforced random walk to countably infinite spaces. However, a closed form for the prior (mixing measure) is lacking and inference in this model is challenging. In this work, we assume countably infinite state space and take the alternative approach of explicitly constructing the prior over the transition matrix. Inference can be done using relatively straighforward Markov Chain Monte Carlo method. We use the resulting reversible Markov chain as the hidden sequence in a Hidden Markov model whose utility we validate on two real world datasets. 



The paper is organized as follows. In Section 2, we briefly provide some background on the Gamma process which is central to our model definition. In Section 3, we describe the process proposed in this manuscript. We discuss its theoretical properties in Section 4 and provide a de Finetti representation for the process in Section 5. The finite version of the model and its HMM extension appear in Sections 6 and 7 and inference, performed via a Gibbs sampler, is described in Section 8. In Section 9 we study our model's performance on real datasets and in Section 10 we conclude our work and provide a short discussion about future directions.

\section{The Gamma Process}
To facilitate understanding, we briefly review the Gamma process $\Gamma P(\alpha_0, \mu_0)$ over a space $\mathcal{X}$. A realization $G_0 \sim \Gamma P $ is a positive measure on the space $\mathcal{X}$, which can be represented as a countable weighted sum of atoms.  Each atom $i$ at $x_i\in\mathcal{X}$ has corresponding weight $w_i \in (0, \infty)$. The atoms and weights are distributed according to a Poisson process over the product space $\mathcal{X} \times [0, \infty)$ with intensity measure
\begin{align}\label{eq:nu}
\nu(dw,dx) = \rho(dw) \mu_0(dx)  = a_0 w^{-1}e^{-a_0 w}dw ~\mu_0(dx) .
\end{align}
where $\mu_0$ is the base measure and $\alpha_0$ is concentration parameter. $\nu$ is known as the L\'evy measure of the gamma process, and because of this representation the gamma process is a L\'evy process. 
In this paper, we assume that the base measure is the Lebesgue measure. 
A sample from this Poisson process will yield a \textit{countably infinite} collection of atoms $\{x_i, w_i\}_{i=1}^{\infty}$ since $\int_{\mathcal{X} \times [0, \infty)} \nu(dw,dx) = \infty$. We assume $\mu_0$ is diffuse (non-atomic) and so, we can write:
 \begin{align}\label{eq:G0}
G_0 := \sum_{i=1}^\infty w_i \delta_{x_i} \sim \Gamma P(\alpha_0, \mu_0)
\end{align}
Intuitively, $G_0(A)$ sums up the values of $w_i$ with $x_i \in A$. It can be shown that the distribution of $G_0(A)$, where $A \subseteq \mathcal{X}$, is $\text{Gamma}(\alpha_0\mu_0(A), \alpha_0)$, hence the name of the process. The gamma process is a completely random measure \citep{King67} and as such for any disjoint and measurable partition $A_1, \dots, A_n$ of $\mathcal{X}$ the random variables $\{G_0(A_1), \dots, G_0(A_n)\}$ are mutually independent gamma variables.

\section{Model Description}
Given a measurable space $\{\mathcal{X}\text{, }\mathcal{F}\}$, with a set $\mathcal{X}$ and a $\sigma$-algebra $\mathcal{F}$ of subsets of $\mathcal{X}$, our aim is to construct a model, a sample from which will give rise to a reversible Markov chain of states $ X_1, \dots,  X_t ,\dots, X_T$. At each time point $t$ the chain is at a state $x$ denoted as $X_t = x$. We require that the set $\mathcal{S}:= \{ x_i \in \mathcal{X},~ i \in \mathbb{N}\}$ is countable. We construct the prior by deploying the gamma process in a hierarchical fashion; we use a gamma process to sample the states and given these states, we construct the transition matrix by sampling from another gamma process. 

More carefully, let $\Gamma\text{P}(\alpha_0, \mu_0)$ be a gamma process on $\mathcal{X}$, with concentration parameter $\alpha_0$ and base measure $\mu_0$, as given by Equation \ref{eq:nu}. A sample $G_0$ from this process corresponds to the set of atoms $\mathcal{S} = \{ x_i \in \mathcal{X}, ~i\in \mathbb{N}\}$ and their associated weights $\{w_i\}_{i=1}^{\infty}$ as in Equation~\ref{eq:G0}. Note that by construction the cardinality of the set $\{x_i\}_{i=1}^{\infty}$ is countably infinite and there is a one-to-one mapping of the atoms in $\mathcal{S}$ to the set of natural numbers $\mathbb{N}$. We define a new gamma process $\Gamma\text{P}(\alpha, \mu)$ on the product space $\mathcal{S} \times \mathcal{S}$, with concentration parameter $\alpha$ and atomic base measure
\begin{align}\label{eq:mu}
\mu(x_i,x_j) = G_0(x_i)G_0(x_j)
\end{align}
where $\mathcal{S}$ is the support of $G_0$. The base measure $\mu$ is atomic and as such, assigns non-zero mass on atoms on the product space $\mathcal{S} \times \mathcal{S}$.
Since $G_0$ is discrete a.s., $G$ will also be discrete so we can write
\begin{align}
G = \sum_i \sum_j J_{ij} \delta_{x_ix_j},
\end{align}
where, from the definition of the Gamma process with fixed points of discontinuity $(x_i,x_j)$, we have
\begin{align}\label{eq:J}
J_{ij}|G_0 \sim \text{Gamma}(\alpha \mu(x_i,x_j), \alpha) = \text{Gamma}(\alpha w_i w_j, \alpha)
\end{align}
where $\alpha w_i w_j$ is the shape and $\alpha$ the rate of the gamma distribution. To avoid notation overload, we also use $J$ to represent the weight matrix which when normalised per row, gives the transition matrix $P$ such that $P_{ij} = P(x_i, x_j) = \frac{J_{ij}}{\sum_{\kappa} J_{i\kappa}}$ is the probability of transitioning to state $x_j$ given that the chain is in state $x_i$ currently. The transition matrix $P$ is stochastic, that is, its entries are all non-negative and $\sum \limits_{j :x_j \in \mathcal{S}} P_{ij} = 1$, for all $x_i \in \mathcal{S}$. By the additive property of the gamma process, each row $J_j$ in the weight matrix is still a sample from a gamma process in the restricted space 
$\{x_j\} \times \mathcal{S}$ with base measure $\mu$, so
\begin{align}
G(\{x_j\},\cdot) = \sum_i J_{ji} \delta_{x_i}
\end{align}

To generate the sequence $X_1, \dots, X_n$, we draw an initial state $X_1 \sim \tilde{G}_0$, where $\tilde{G}_0$ is the normalised random measure derived from $G_0$, i.e. $\tilde{G_0} = G_0/G_0(\mathcal{X})$ and sample the transition $X_{n-1} \rightarrow X_{n}$, as follows:
\begin{align}
X_n | X_{n-1}, G \sim \frac{G(X_{n-1},\cdot)}{G(X_{n-1},\mathcal{S})}=P(X_{n-1}, \cdot)
\end{align}

The process can be thought as a weighted random walk on a graph, with vertex set $\mathcal{S}$ and edge set $\{(x, y) \in \mathcal{X}^2 ; J_{ij}>0\}$. $J: \mathcal{S} \times \mathcal{S} \rightarrow (0, \infty)$ is a function that puts non-negative weight to each edge in the graph. 

The above random walk has not yet yielded a \emph{reversible} Markov chain. To achieve this reversibility we modify \eqref{eq:J} so that $J_{ij}$ is symmetric, i.e. 
\begin{align} \label{eq:symmetry}
J_{ij} = J_{ji} | G_0 \sim  \text{Gamma}(\alpha w_iw_j, \alpha)
\end{align}
Note that the function $J$ is now symmetric and the new $G$ defined using symmetric $J$ is no longer a draw from a completely random measure because of the dependency induced by this symmetry. However, each row is still a draw from a completely random measure. The resulting transition matrix $P$ is a sample from the prior, the construction of which was just described. 

\begin{figure}
\centering
\subfigure[SHGP]{\includegraphics[height=2in]{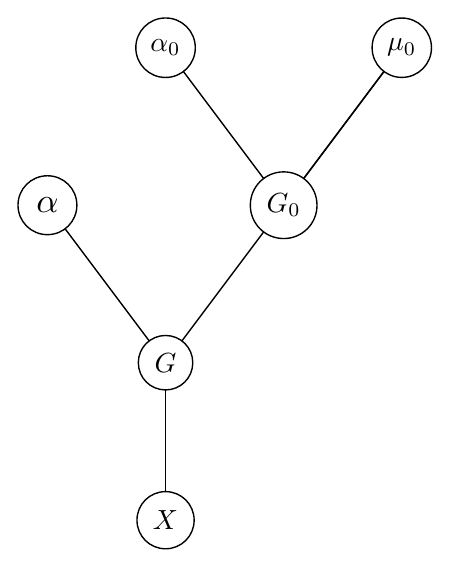}}
\subfigure[SHGP - HMM]{\includegraphics[height=2in]{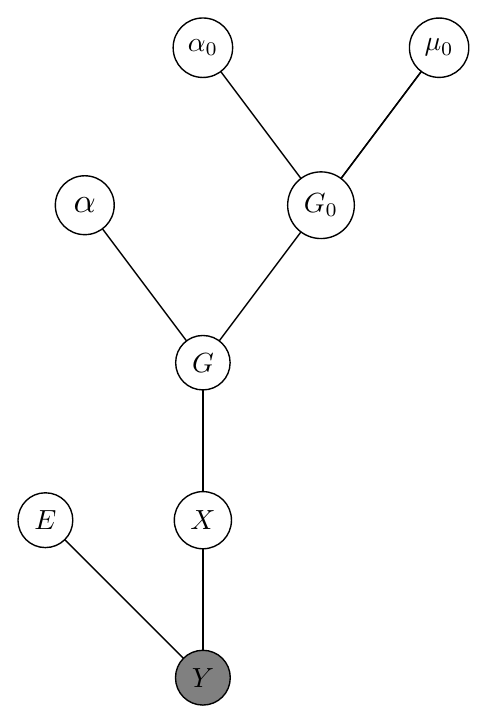}}
\vspace{0.1in}
\caption{(a)Graphical model for SHGP and (b) SHGP as part of an HMM where the time series $X$ and $Y$ are represented as single nodes.}
\vspace{0.1in}
\label{fig:graph_shgp}
\end{figure}

We note here that the choice of the shape value for each $J_{ij}$ weight might not be restricted to the product of the corresponding $w$'s. Depending on the dataset at hand, the choice might vary. We call the proposed model Symmetric Hierarchical Gamma Process and use the acronym SHGP. A graphical representation of the model is presented in Figure \ref{fig:graph_shgp}(a).

\paragraph{Relation to Hierarchical Dirichlet process}
The construction of the proposed SHGP prior closely relates to the Hierarchical Dirichlet process \cite{TehJorBea2006}. Both processes use random measures in a hierarchical way: the HDP uses the Dirichlet, while the SHGP the Gamma process as seen in Table \ref{tab:hdp-shgp}, where $J_j$ refers to the $j$-th row of the weight matrix. Moreover, both processes when used for infinite Hidden Markov models \cite{beal03}, put a prior on the transition matrix $P$ but in a different fashion; the HDP directly defines a prior over $P$, while the SHGP puts a prior on the weight matrix $J$, the per-row normalisation of which gives the transition matrix. As such, the SHGP allows for a direct treatment of the weigths, the symmetry in Equation \ref{eq:symmetry} is imposed and reversibility arises.

\paragraph{Relation to Hierarchical Gamma process}
Interesting is the relation of the SHGP to the simple Hierarchical Gamma process (HGP) also seen in Table \ref{tab:hdp-shgp}. Both processes use the Gamma process in a hierarchical way. The HGP does not assume symmetry in the weights but this can be easily imposed. However, the random variable $J_{ij}$ is sampled from Gamma distributions with different shape parameters. More specifically
\begin{align}\label{eq:hgp-shgp}
J_{ij} = J_{ji} | G_0 \sim  \text{Gamma}(\tilde{\alpha} w_i) \text{~~ for the HGP}\nonumber \\
J_{ij} = J_{ji} | G_0 \sim  \text{Gamma}(\alpha w_i w_j) \text{~~ for the SHGP}
\end{align}
As seen in Equation \ref{eq:hgp-shgp}, in SHGP the base weights of both the nodes $i$ and $j$ contribute to the edge weight $J_{ij}$, as opposed to the HGP where only one of the base weights influences the shape. As already stated earlier in this Section, this is a modelling choice that depends on whether or not contribution of both nodes is desired. More details about the relation amongst SHGP, HGP and HDP can be found in the supplementary material.

\begin{table}
\caption{HDP, HGP and SHGP} 
\label{tab:hdp-shgp}
\begin{center}
\begin{sc}
\begin{tabular}{ccc}
\hline
HDP & HGP & SHGP\\\hline \\
$G_0' \sim \text{DP}(\alpha_0 \mu_0)$  & $ G_0 \sim \Gamma P(\alpha_0, \mu_0)$  &	$ G_0 \sim \Gamma P(\alpha_0 ,\mu_0)$ \\
$J_j \sim \text{DP}(\alpha' G'_0)$  &  $ \tilde{J_j} \sim \Gamma P(\tilde{\alpha}, G_0)$ &	$ J_j \sim \Gamma P(\alpha w_j, G_0)$ \\
\hline
\end{tabular}
\end{sc}
\end{center}
\end{table}

\section{Theoretical Properties}
In this section, we describe important theoretical properties of the induced Markov chain given the sample from the SHGP process. The theory used, is the theory applied on Markov chains on countably infinite space since the induced Markov chain falls in this category. 

\paragraph{Reversibility}
In order to prove that the induced Markov chain is reversible, it is sufficient to prove that detailed balance holds, that is
\begin{equation}
\pi(x_i)P_{ij} =  \pi(x_j)P_{ji} 
\end{equation}
where $\pi$ is the probability defined by $\pi(x_i) = \frac{\sum_{\kappa} J_{i\kappa}}{ \sum_l \sum_{\kappa} J_{l\kappa} }$ and ${P}$ is the stochastic transition matrix induced by the rows of the symmetrised $G$.
\paragraph{Proof}:\\
We have
\begin{align}
 \pi(x_i)P_{ij} &= \frac{\sum_{\kappa} J_{i \kappa}}{ \sum_l \sum_{\kappa} J_{l\kappa} }  \frac{J_{ij}}{\sum_{\kappa} J_{i\kappa}}  = \frac{J_{ij}}{\sum_l \sum_{\kappa} J_{l\kappa}}\nonumber \\
 &=\frac{J_{ji}}{\sum_l \sum_{\kappa} J_{l\kappa}}  =  \frac{\sum_\kappa J_{j \kappa}}{\sum_l \sum_{\kappa} J_{l\kappa} }  \frac{J_{ji}}{\sum_{\kappa} J_{j\kappa}} \nonumber \\
 &= \pi(x_j)P_{ji},
\end{align}
as a result of \eqref{eq:symmetry}.  As a straighforward corollary,  $\pi$ is the invariant measure of the chain.

\paragraph{Is the normalization constant per row finite?}
When defining the transition matrix $P$, it is crucial that the sum of each row in the weight matrix $J$ is almost surely (a.s.) finite, since this ensures that the normalization is a well-defined operation. In other words, we want to ensure that for every row $j$ in the weight matrix $\big|\sum_{i}J_{ji}\big| < \infty$ holds a.s. To start with, the sum $\big|\sum_{i}w_{i}\big|$ converges a.s., that is
\begin{equation}\label{eq:norm_o}
\big|\sum_{i}w_{i}\big| < \infty, \quad a.s.
\end{equation}
if the well-known condition on the Levy measure $\rho(dw)$ that
\begin{equation}\label{eq:cond}
\int_{R^+}(1-e^{-w})\rho(dw) < \infty,
\end{equation}
holds. For a Gamma process where $\rho(dw) = a_0 w^{-1}e^{-a_0 w}$ it is easy to prove that the above condition holds and as such the sum in \eqref{eq:norm_o} converges. Since the weights $w_{i}$ are defined over the space $[0, \infty)$, we ensure that $w_i\geq0$ always. Consequently, we can drop the absolute value notation and simply write $\sum_{i}w_{i}<\infty$. Moreover, since the measure over $w$ is continuous, $P(w_i=0)=0$ for $\forall i$ and $\sum_{i}w_{i}>0$ a.s.

The sum in each row $i$ in the weight matrix is $\sum_{j}J_{ij}$. Each element $J_{ij}$ of this sum is a gamma distributed variable sampled from the gamma distribution $\text{Gamma}(\alpha w_i w_j, \alpha)$. Recall here that the variables $J_{ij}$ and $J_{ji}$ are being sampled from the same Gamma distribution. This, along with the property that the sum of gamma distributed variables with the same rate parameter is a gamma distributed variable with the shape equal to the sum of the shape parameters of the individual gamma variables and the same rate gives the following marginally
\begin{equation}
\sum_i J_{ji} \sim \text{Gamma}(\alpha w_j \sum_i w_i, \alpha)
\end{equation}
Since we have ensured that $0<\sum_i w_i < \infty$ a.s., the sample $\sum_i J_{ji}$ is finite a.s., ensuring that the normalization for every row in the weight matrix is a well-defined operation.

\paragraph{Irreducibility and aperiodicity.}
A  Markov chain is irreducible if it is possible to get from any state to any other state in a finite number of steps with positive probability. In other words, when a Markov chain is irreducible, the sample path (the state sequence) cannot get trapped in smaller subsets of the state space. That is, for any two states $x_i, x_j \in \mathcal{X}$ there exists an integer $t$, such that $P_{ij}^t>0$. It is easy to see that the proposed generative process produces an irreducible Markov chain almost surely. Looking at the weight matrix, we see that the elements are gamma distributed variables with support $J_{ij}\in (0, \infty)$, and thus are positive almost surely. This, along with the existence (and finiteness) of the normalization constant shows that the probability of moving from one state to any other in one step is always positive and the chain is irreducible. Let $T_{ii}:=\{t\geq1 : P^{t}_{ii} >0\}$ be the set of times when it is possible for the chain to return to starting state $X_i$. The period of the 
state $X_i$ is defined to be the greatest common divisor of $T_{ii}$. For an irreducible chain, the period of it is defined to be the period which is common for all the states. We note that the transition matrix is strictly positive and as a result the chain can be in any state in one step. This implies that all the states have period 1 and the chain is aperiodic. 
    
\paragraph{Convergence}
We showed that the generated Markov chain has an invariant probability distribution $\pi$.
A state $x_i$ is positive recurrent if the expected amount of time to return to state $i$ given that the chain started in state $x_i$ has finite first moment that is, $E(\tau_{ii}) < \infty$, where $\tau_{ij}:= \text{min}\{n \geq 1 : X_n=x_j | X_0 = x_i\}$ is the time (after time 0) until reaching state $x_j$ given $X_0 = x_i$.
 An irreducible Markov chain with transition matrix $P$ is positive recurrent if and only if there exists a probability distribution $\pi$ on $\mathcal{X}$ such that $\pi=\pi P$ [Theorem 21.12, \citep{Levin06}]. As such, the generated Markov chain $\{S_1, S_2\dots\}$ is positive recurrent. Irreducibility, aperiodicity and positive recurrence ensure that the invariant distribution $\pi$ is unique and for all $x \in \mathcal{X}$ [Theorem 21.14, \citep{Levin06}] ,
\begin{equation}\label{eq:conv}
\lim_{t \to +\infty} ||P^{t}(x, \cdot) - \pi ||_{TV} = 0
\end{equation}
where $TV$ denotes the total variation distance between the two distributions. Equation \eqref{eq:conv} describes the convergence of the chain as $t \to +\infty$ and states that every row in the transition matrix $P^t$ converges to the stationary distribution $\pi$ eventually. In other words, the invariant distribution $\pi$ is also the limit distribution of the chain. 

\section{de Finetti Representation}
\citet{diaconis80} defined a type of exchangeability for Markov chains, known as Markov exchangeability and it is defined for sequences $X_1, X_2, \dots$ in a countable space $\mathcal{X}$ as follows:

\begin{definition}\label{mxchng}
 A process on a countable space $\mathcal{X}$ is Markov exchangeable if the probability of observing a path $X_1, \dots, X_n$ is only a function of $X_1$ and the transition counts $C(x, y) := |\{ X_1=x, X_{i+1}=y; 1\leq i<n \}|$ for all $x ,y \in \mathcal{X}$.
\end{definition}
In other words, a sequence is Markov exchangeable if the initial state $X_1$ and the transition counts are sufficient statistics. Intuitively, this means that two
different state sequences are equiprobable under the joint distribution, if they begin with the same value and preserve the transition counts between unique values. They also proved the following
\begin{thm}[Diaconis and Freedman, 1980]\label{diaconis}
A process is Markov exchangeable and returns to every state visited infinitely often (recurrent), if and only if it is a mixture of recurrent Markov chains. 
\end{thm}
In the previous Sections, we defined a prior over transition matrices using a hierarchy of gamma processes. We also proved that the induced Markov chains (given the transition matrix sampled from the prior) are recurrent. The use of the prior let us write the state sequence as a mixture of recurrent Markov chains and using Theorem \ref{diaconis} we can state that the sequence $\{X_n\}$ generated by the proposed process and defined on a countably infinite space $\mathcal{S}$ is Markov exchangeable and recurrent.  

\begin{proposition}\label{prop}
For some measure $\varphi$ on $\mathcal{S} \times \mathcal{P}$, where $\mathcal{P}$ is the space of stochastic matrices on $\mathcal{S} \times \mathcal{S}$, the distribution of $(X_i)_{i \in \mathbb{N}}$, can be represented as 
\begin{equation}\label{eq:finetti}
p(X_1, \dots, X_n) = \int_{\mathcal{P}} \prod_{i=1}^{n-1}P(X_i, X_{i+1})\varphi(X_1, dP)
\end{equation}
\end{proposition}
Equation \eqref{eq:finetti} shows the de Finetti representation of the proposed process. The de Finetti measure is the distribution $\varphi$ over the product of the space $S$ and the space of stochastic matrices $\mathcal{P}$.

\section{Finite Model}

The inference simplifies considerably if we consider the finite state model which gives the countably infinite state model in the limit. More carefully, we assume that we have a finite number of states $K$ and we prove that as $K \rightarrow \infty$, the model converges in distribution to the countably infinite model. 

The infinite divisibility property of the gamma process $G_0$ on $\mathcal{X}$ states that for each $K=1 ,2, \dots$ there exists a sequence of i.i.d. random variables $G_0(A_{1}) + \dots + G_0(A_{K})$ such that 
\begin{equation}\label{eq:infdiv}
G_0(\mathcal{X}) \stackrel{d}{=} G_0(A_{1}) + \dots + G_0(A_{K})
\end{equation}
where $\stackrel{d}{=}$ is equality in distribution. Due to the additive property of the Gamma distribution, for any finite, disjoint and measurable partition $A_1, \dots, A_K$ of $\mathcal{X}$ such that $\mathcal{X} = \bigcup_{i=1}^{K}A_i$, the variable $G_0(\mathcal{X})$ with law $\text{Gamma}(\alpha_0\mu_0(\mathcal{X}), \alpha_0)$ can be written as the sum of $K$ Gamma distributed variables each one following the law $\text{Gamma}(\alpha_0 \mu_0(A_j) , \alpha_0)$, that is $G_0(\mathcal{X}) = \sum_{j=1}^{K}G_0(A_j)$. The additive property of $\mu_0$ ensures that $\mu_0({\mathcal{X}})  = \sum_{i=1}^{K}\mu_0(A_i)$ and as such the shape parameter of the Gamma distribution of $G_0(\mathcal{X})$ will be equal to the $\alpha_0\sum_{j=1}^K \mu_0(A_j)$. As $K \rightarrow \infty$ we recover the infinite case and Equation~\eqref{eq:infdiv} holds. For simplicity, we assume that $\mu_0(A_j) = \frac{\mu_0(\mathcal{X})}{K}$. 

By restricting the process to the finite case, we facilitate inference without compromising the properties of the model since $K$ can always be chosen sufficiently large.  Putting everyting together, the generative process in the finite case is as follows:
\begin{align}\label{eq:genpro}
G_0 &= \sum_{i=1}^K w_i \delta_{x_i} \nonumber \\ 
w_i &\sim \text{Gamma}(\alpha_0 \mu_0(x_i), \alpha_0) \nonumber \\
G &= \sum_{i=1}^K \sum_{j=1}^K J_{ij} \delta_{x_i, x_j} \nonumber \\ 
J_{ij} = J_{ji} &\sim \text{Gamma}(\alpha w_i w_j, \alpha)
\end{align}

\section{The SHGP Hidden Markov model}

In typical sequential data analysis we are more interested in using a Markov chain as the hidden state sequence in a Hidden Markov model (HMM) rather than viewing $X$ as observations themselves. This allows a broad range of data types to be modelled: the example we will demonstrate here include univariate continuous and multivariate count data. Thus we use the SHGP to construct a Hidden Markov model. Consider a sequence of observations $\{ Y_t \in \mathcal{Y} : t=1,\dots,T \}$ which we will assume to be independent conditioned on the latent state sequence $X$. For simplicity consider $X_t \in \{ 1, \dots, K\}$ under the finite SHGP, and a parametric family of observation models $F(\cdot|\theta)$, then
\[
 Y_t | X_t, \theta \sim^{iid} F(\cdot|\theta_{X_t})
\]
where $\{ \theta_k, k=1,\cdots,K \}$ are the state emission parameters. In the case of multinomial outputs $\theta_k$ is a probability vector, the concatenation of which is known as the emission matrix. The SHGP gives the prior over the hidden state sequence as shown in Figure \ref{fig:graph_shgp}(b). We present multinomial, Poisson and Gaussian observation models $F(.)$, the details for which are given in the supplementary material.  

\section{Inference}
As with many other Bayesian models, exact inference is intractable so we employ Markov Chain Monte Carlo (MCMC) and using an iterative process we achieve posterior inference over the latent variables of the model as seen in Figure \ref{fig:graph_shgp}(b). A detailed description of the sampling steps is provided in the supplementary material. The sampler iterates as follows:

\paragraph{Sampling the concentration parameters, $\alpha_0$ and $\alpha$.}
We used slice sampling by \citep{Neal2003} to infer the parameters $\alpha0$ and $\alpha$ using Gamma priors $\alpha_0 \sim \text{Gamma}(s_0, r_0)$ and $\alpha \sim \text{Gamma}(s, r)$, where $\{s_0, r_0\}$ and $\{s, r\}$ are the pairs of shape and rate parameters for $\alpha_0$ and $\alpha$ respectively. 

\paragraph{Sampling the weight vector, $G_0$} The vector $G_0$ is the vector of the base weights $G_0 = [w_1, \dots, w_K]$ in the corresponding random measure $G_0 = \sum_k w_k \delta_{x_k}$.
We used slice sampler to sample each weight $w_k$.

\paragraph{Sampling the weight matrix, $J$} The weight matrix $J$ contains the edge weights $\{J_{ij}\}$. We used hybrid Monte Carlo \citep{Neal2011} to sample the elements of the matrix at once instead of sampling each element at a time using slice sampling. In the reversible case, only $K(K+1)/2$ are sampled because of the symmetry in $G$. We also show results using NUTS~\citep{hoffman2011no} although our results suggest this gives similar performance to HMC in this setting. 

\paragraph{Sampling the state sequence $\mathbf{X}$}
We use the forward-backward algorithm \citep{Scot02} to sample the latent state sequence $X$ given the current state of all other variables in the model. This is a dynamic programming algorithm that efficiently computes the state posteriors over all the hidden state variables $X_t$. 

\paragraph{Sampling the emission matrix $E$}
The posterior over the emission matrix is $$p(E | Y, X) \propto p(Y| E, X)p(E)$$
The explicit form of the posterior depends on the output, the observed $Y$, that is multinomial, Poisson or Gaussian. In all cases, due to conjugacy, the emission matrix is sampled exactly.

\section{Experiments}
In this section we evaluate the SHGP by running SHGP Hidden Markov model on two real world datasets. The datasets are especially chosen such that the underlying systems are reversible. For completion, we also ran SHGP assuming non-reversibility by not imposing symmetry in the inferred weight matrix $J$. Moreover, we compare SHGP to the infinite HMM which learns a transition matrix for the hidden state sequence and does not account for reversibility. For the iHMM we use the beam sampler~\citep{VanGael08}. 

\paragraph{Prediction} A principled way to evaluate a generative model
is by its ability to predict missing data values given some observations. 
For SHGP we collect $M$ samples from the posterior $\{ \{E^{(1)}, X^{(1)}\}, \dots, \{E^{(M)}, X^{(M)}\} \}$ and estimate the predictive distribution of a missing entry in the dataset $Y$ as the average of the predictive distributions 
for each of the collected samples. For the experiments we ran, we used two different likelihoods, a Poisson and a Gaussian. For the Poisson model the approximate predictive distribution is 
\begin{equation*}
P(Y_{lt}|Y_{train}) \approx \frac{1}{M} \sum_{m=1}^{M} \frac{E(X^{(m)}_t, l)^{Y_t(l)} - e^{E(X^{(m)}_t,l)}}{Y_t(l)!}\text{,}
\end{equation*}
while for the Gaussian is
\begin{equation*}
P(Y_{t}|Y_{train}) \approx  \frac{1}{M}  \sum_{m=1}^{M}  \frac{1}{E(X^{(m)}_t,2) \sqrt{2\pi}} e^{- \frac{(Y_t - E(X^{(m)}_t, 1)^2)}{2E(X^{(m)}_t,2)^2}}
\end{equation*}
The supplementary material provides a detailed description of the likelihood models.

\subsection{ChIP-seq epigenetic marks}

For this experiment we used ChIP-seq (chromatin immunoprecipitation sequencing) data, representing histone modifications and transcription factor binding in human neural crest cell lines (see \citet{park2009chip} for a nice review). ChIP-seq is a method to identify the sites in a DNA sequence where specific proteins are bound. The workflow of ChIP-seq is: 1) DNA is extracted from cells, 2) the proteins of interest (POI) and DNA are chemically bound (``cross-linked''), 3) the DNA is fragmented using sonification, 4) an appropriate antibody is used to filter out the DNA fragment bound to the POI using immunoprecipitation, 5) the POI is  removed from the DNA, 6) the DNA is sequenced. The reads are finally mapped to a known reference sequence. Note that reversibility is reasonable here because although individual genes have direction, the genome as a whole has no particular direction. 

\begin{figure}
\centering
\subfigure{\includegraphics[width=\columnwidth]{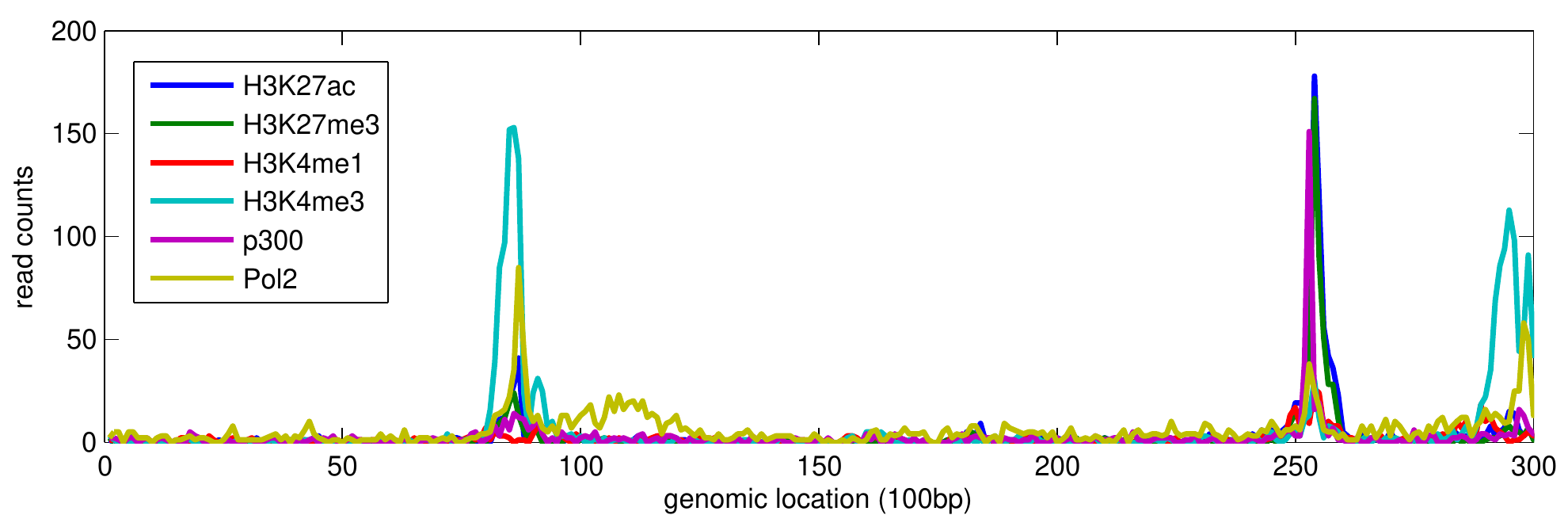}} 
\caption{ChipSeq data for a small region of chromosome 1. The H... markers are histones with various chemical modifications, PolII is RNA polymerase II and p300 is a transcription factor. }
\label{fig:chipseq}
\end{figure}

The resulting observed sequence $Y_{lt}$ is a $L \times T$ matrix of counts, representing how many reads for POI $l$ mapped to bin $t$, where a bin is a 100bp region of the genome (different size bins could be used, but 100bp is commonplace).  A small section of our full $L=6$ by $T=20000$ dataset $Y$, of length $300$, along with the identifiers of the POIs is shown in Figure \ref{fig:chipseq}. 

\begin{figure*}
\label{fig:emChip}
\centering
\subfigure{\includegraphics[width=1.5\columnwidth]{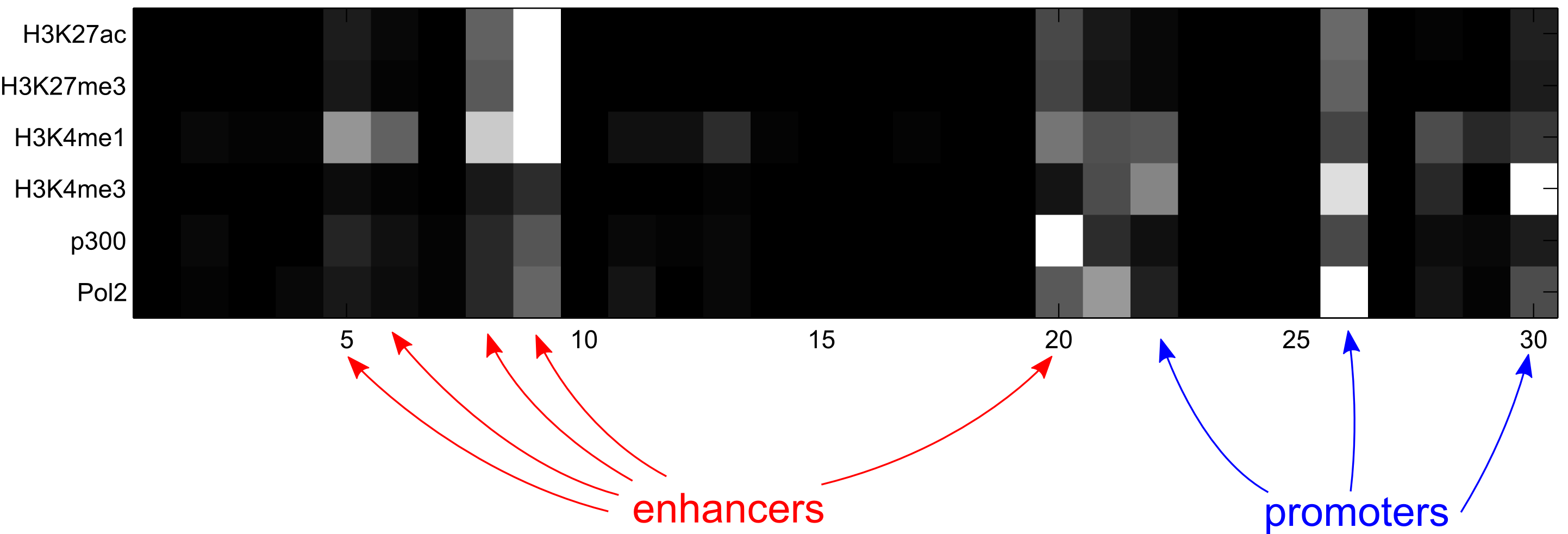}}
\caption{Learnt emission matrix $L \times K$ for ChIP-seq dataset. Element $E_{lk}$ is the Poisson rate parameter for protein $l$ in state $k$. Brighter indicates higher values. Here we associate the states learnt in an supervised manner with known functional regulatory elements, see e.g. \citet{rada2010unique}}.
\end{figure*}

ChiSeq results presented in Table \ref{table:chipR}. We ran 10 repeats, each time holding out different $20\%$ of the data $Y$ and using different random initilisation. The likelihood model used here is Poisson and the task was to predict the held out values in Y. We see that in terms of predictive performance the reversible SHGP-HMM outperforms both the non-reversible version of the model and the iHMM trained using beam sampling. The ``emission'' matrix, the $L$ by $K$ matrix of Poisson rates is shown in Figure \ref{fig:emChip} where we identify expected states known as enhancers and promoters based on their activity levels for the different markers (POIs). 

\begin{table*}
\caption{ChipSeq results for 10 runs using different hold out patterns. We used a truncation level of $K=20$, $1000$ iterations and a burnin of $700$.  }
\label{table:chipR}
\begin{center}
\begin{tabular}{l l c c c c }
\hline
Model& Alogirthm & Train error   & Test error & Train log likelihood & Test log likelihood  \\ \hline
Reversible & HMC & $\mathbf{0.9122 \pm 0.0032 }$ &  $\mathbf{1.1158 \pm 0.0097}$ & $\mathbf{-1.0488\pm 0.0009}$ & $\mathbf{-3.2422 \pm 0.0023}$\\
Non-reversible  & & $0.9127 \pm 0.0033$ & $1.1167 \pm 0.0095$ & $-1.0494 \pm 0.0009$ & $-3.2478 \pm 0.0022$  \\
 iHMM & Beam Sampler & $0.9383 \pm 0.0061$ & $1.1365 \pm 0.0107$ &  $-1.0727 \pm 0.0041$ & $-3.3047 \pm 0.0027$ \\ \hline
\end{tabular}
\end{center}
\label{tab:nips}
\end{table*}

\subsection{Single ion channel recordings}
\begin{figure*}
\centering
\subfigure{\includegraphics[width=1.7\columnwidth]{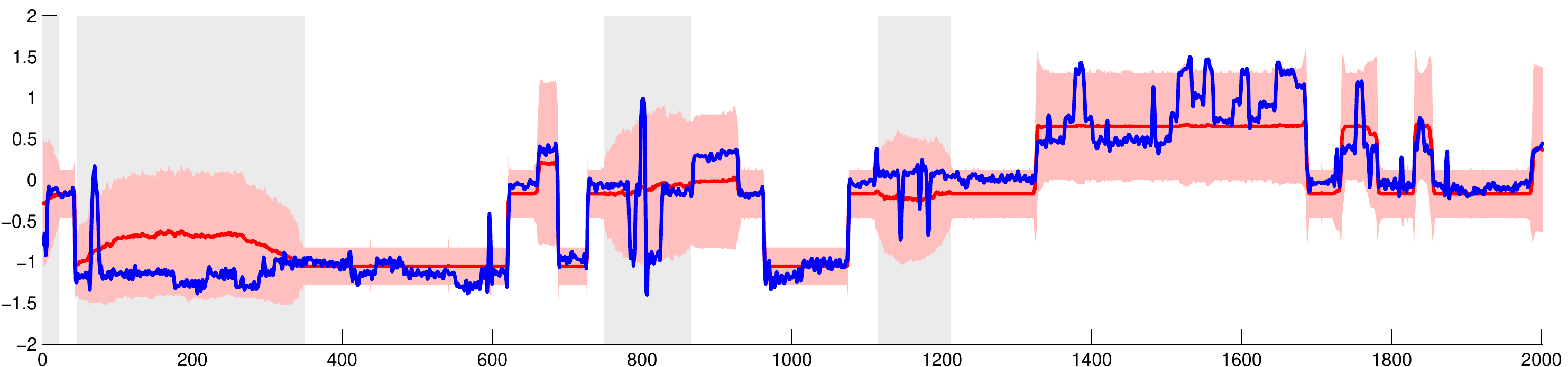}} 
\caption{Ion channel recording (blue) with predictive distribution mean (red) and one standard deviation (pink region) and missing regions used for assessing predictive performance (grey). The predictive variance here includes the variance of the state observation model and uncertainty over the state, calculated by averaging over multiple hidden state samples. As expected the predictive variance in the missing regions is increased and mostly covers the true signal, suggesting the model is well calibrated. } 
\label{fig:ion_plot}
\end{figure*}

Patch clamp recordings are a well established experimental method to measure conformational changes in ion channels, proteins embedded in lipid membranes of cells (such as the cell surface membrane), which control the flow of chemicals such as neurotransmitters across the membrane. These changes are accompanied by changes in electrical potential which can be measured. HMMs have been used to analyse these recordings for many years~\citep{becker1994analysing}, but have usually ignored the prior knowledge that the underlying physical system has time reversible dynamics. We incorporate this information using the SHGP-HMM, analysing a 1MHz recording from the state-of-the-art method of \citep{rosenstein2013single} of a single alamethicin channel. We subsample this time series by a factor of 100 to obtain a $T=10,000$, 10KHz recording, which we log transform and normalise. A small segment of the recording, along with the fitted SHGP-HMM is shown in Figure~\ref{fig:ion_plot}. Grey regions represent aritificial missingness used to test the predictive performance of the models, as shown in Table \ref{table:ionR}. Here we see that the reversible version of SHGP-HMM outperforms both the non-reversible version and the iHMM using the beam sampler, showing the advantage of using the prior knowledge that the transition matrix should be reversible. More specifically, the reversible model performs slightly better in terms of test error than the non-reversible model, although this difference is not quite significant based on paired t-test ($p=0.08$). In terms of test log likelihood the reversible version of the model does perform significantly better however. The use of HMC or NUTS does not significantly impact the results in this case. The iHMM using the beam sampler does significantly worse in terms of train and test error, but not significantly better than the non-reversible HGP-HMM. 

\begin{figure}
\centering
\subfigure{\includegraphics[width=.7\columnwidth]{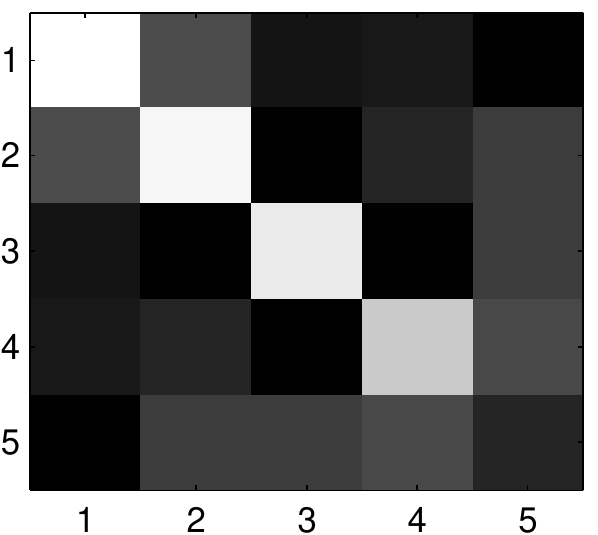}} 
\caption{Learnt weight matrix $J$ for the ion channel recording. The states depicted are those $i$ with $J_{i.} /J_{..}  > 0.01$ where $.$ denotes summation. The states are ordered by the mean of their Gaussian observation model. Transitions between states $1$ and $2$ with the lowest current levels are the most common, followed by between states $2$ and $4$.}
\label{fig:ionR}
\end{figure}

\begin{figure}
\centering
\subfigure{\includegraphics[width=\columnwidth]{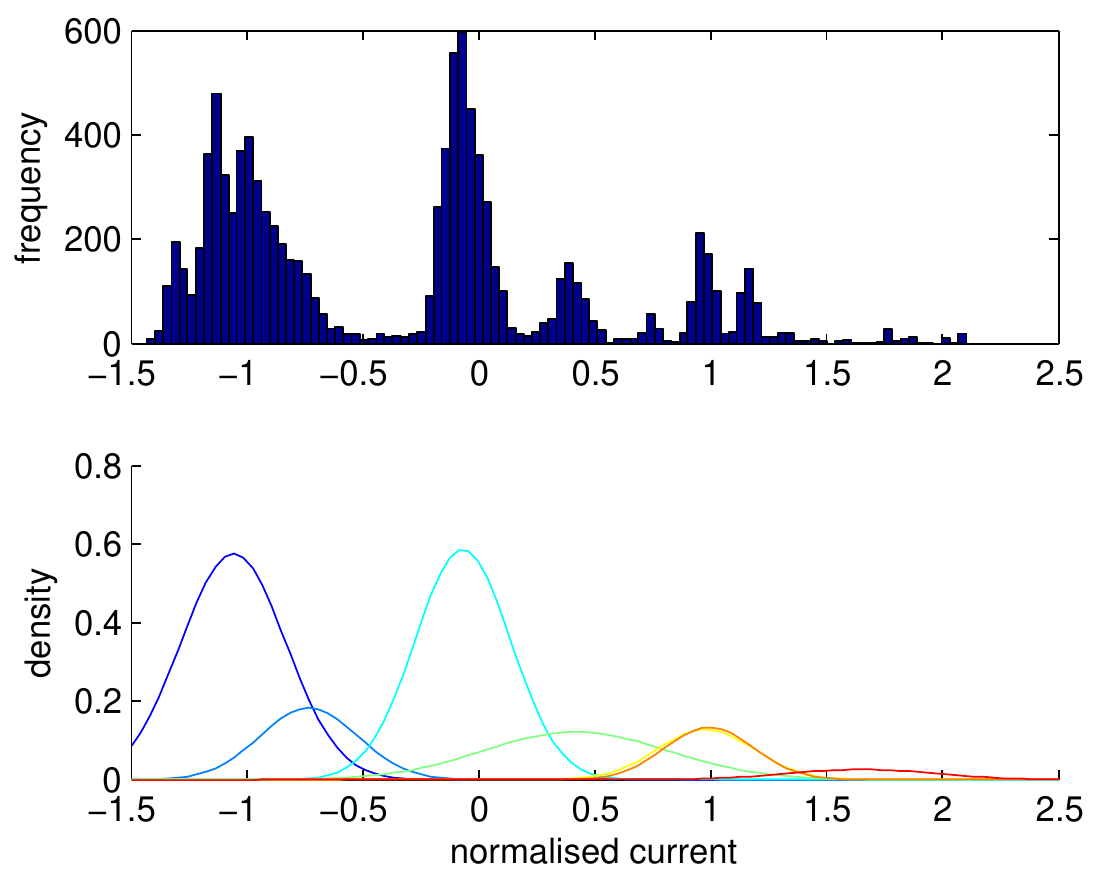}} 
\caption{Clusters found by the sHGP-HMM for the ion channel dataset, shown relative to a histogram of levels across the recording. The smaller clusters at higher currents are often merged in the model. }
\label{fig:ionClusters}
\end{figure}

SHGP-HMM typically uses $5$ to $7$ states for this dataset. A typical sample of $J$ is shown in Figure \ref{fig:ionR} and the observation models for each state are illustrated in Figure \ref{fig:ionClusters}. Comparing the histogram of currents to the learnt observation models we see that some of the less common high energy states are blurred into one, which could possibly be alleviated by more careful selection of priors. An additional difficulty of ion channel recordings is that the current level for a particular state tends to drift over time, which is not a characteristic currently supported by our model. 

\begin{table*}
\caption{Ion channel results across 10 different random hold out patterns. For SHGP-HMM we used a truncation of $K=15$, $1000$ iterations and a burnin of $700$. $50$ inner iterations of HMC or NUTS were run per outer iteration.}
\label{table:ionR}
\begin{center}
\begin{tabular}{l l c c c c }
\hline
Model& Alogirthm & Train error   & Test error & Train log likelihood & Test log likelihood  \\ \hline
  Reversible & HMC  & $ 0.023 \pm 0.001 $  & $ 0.030 \pm 0.002 $  & $ \mathbf{ 2.204 \pm 0.055} $  & $ \mathbf{ 2.034 \pm 0.058} $ \\ 
 Non-reversible & HMC  & $ 0.027 \pm 0.007 $  & $ 0.033 \pm 0.007 $  & $ 2.108 \pm 0.084 $  & $ 1.970 \pm 0.078 $ \\ 
 Reversible & NUTS  & $  0.024 \pm 0.004  $  & $ 0.031 \pm 0.003 $  & $\mathbf{  2.190 \pm 0.063} $  & $  2.030 \pm 0.062 $ \\ 
 Non-reversible & NUTS  & $ 0.025 \pm 0.005  $  & $ 0.032 \pm 0.004 $  & $ 2.142 \pm 0.086 $  & $ 1.989 \pm 0.067 $ \\ 
 iHMM & Beam sampler  & $ 0.038 \pm 0.005 $  & $ 0.045 \pm 0.004 $  & $ 2.134 \pm 0.070 $  & $ 2.008 \pm 0.058 $   \\ \hline
\end{tabular}
\end{center}
\end{table*}

\section{Discussion}
Reversibility is a property met in various datasets, especially in ion channel recordings. In this paper, we have introduced a hierarchical non-parametric model, SHGP, which gives rise to reversible Markov chains.  We have used the SHGP to construct a Hidden Markov model allowing a broad range of data types to be modelled. Our experimental results on two different datasets suggest that accounting for reversibility intrinsically in SHGP results in gains in empirical performance compared to non-reversible models. An interesting direction for future work would be to apply the SHGP to MCMC itself: in this settings, the second eigenvalue of the learnt transition matrix could be use as a measure of the mixing perform of the MCMC chain.

\bibliography{ghmm}
\bibliographystyle{apalike}
\end{document}